\documentclass[conference]{IEEEtran}

\usepackage{graphicx}
\usepackage{bm}
\usepackage{amsmath}
\usepackage{amssymb}
\usepackage{esvect}
\usepackage{amsmath}
\usepackage{amssymb}
\usepackage[mathscr]{euscript}
\usepackage{csvsimple}
\usepackage{pgfplotstable}
\usepackage{nicefrac}

  \newcommand{\T}{^\mathsf{T}}
  
  \newcommand{\R}{\mathbb{R}}

  \newcommand{\diag}[1]{\mathrm{diag}(#1)}

  \newcommand{\N}{\mathrm{N}}

  \usepackage{bm}
  \newcommand{\mathbold}[1]{\bm{#1}}
  \newcommand{\mbf}[1]{\mathbf{#1}}
  \newcommand{\vect}[1]{\mbf{#1}}
  \newcommand{\vectb}[1]{\bm{#1}}

\newcommand{\vdelta}[0]{\mathbold{\delta}}

\renewcommand{\mid}[0]{\,|\,}
\newcommand{\vzero}[0]{\mathbold{0}}

\newcommand{\vh}{\mbf{h}}

\newcommand{\vp}{\mbf{p}}
\newcommand{\vq}{\mbf{q}}
\newcommand{\vr}{\mbf{r}}

\newcommand{\vx}{\mbf{x}}

\newcommand{\MA}{\mbf{A}}

\newcommand{\MD}{\mbf{D}}

\newcommand{\MH}{\mbf{H}}
\newcommand{\MI}{\mbf{I}}

\newcommand{\MK}{\mbf{K}}

\newcommand{\MP}{\mbf{P}}
\newcommand{\MQ}{\mbf{Q}}
\newcommand{\MR}{\mbf{R}}

  \newcommand{\eg}{\textit{e.g.}}
  \newcommand{\ie}{\textit{i.e.}}

  \newcommand{\etal}{\textit{et~al.}}

  \usepackage{booktabs}

  \usepackage{tikz,pgfplots}
  \usetikzlibrary{plotmarks,shapes,arrows}
  \pgfplotsset{compat=newest} 
  \pgfkeys{/pgf/number format/.cd,1000 sep={}}

  \newlength\figureheight
  \newlength\figurewidth

  \usepackage[pagebackref=false,breaklinks=true,letterpaper=true,colorlinks,bookmarks=false, citecolor=black, linkcolor=black, urlcolor=black]{hyperref}

  \usetikzlibrary{external}

  \usepackage[capitalize,nameinlink]{cleveref}
  \crefname{section}{Sec.}{Sects.}
  \crefname{proposition}{Prop.}{Props.}
  \crefname{lemma}{Lem.}{Lems.}
  \crefname{model}{Mod.}{Mods.}
  \crefname{appendix}{App.}{Apps.}

  \makeatletter
  \let\NAT@parse\undefined
  \makeatother

  \makeatletter
  \let\NAT@parse\undefined
  \makeatother
  \usepackage[square,numbers,sort&compress]{natbib}

  \makeatletter
  \let\MYcaption\@makecaption
  \makeatother

  \usepackage[font=footnotesize]{subcaption}

  \makeatletter
  \let\@makecaption\MYcaption
  \makeatother

\definecolor{mycolor0}{rgb}{0.2667,0.4471,0.7098}
\definecolor{mycolor1}{rgb}{0.1647,0.6706,0.3804}
\definecolor{mycolor2}{rgb}{0.8275,0.2627,0.3059}
\definecolor{mycolor3}{rgb}{0.5216,0.4392,0.7176}
\definecolor{mycolor4}{rgb}{0.8118,0.7255,0.4118}
\definecolor{mycolor5}{rgb}{0.2745,0.7176,0.8157}

\definecolor{mylcolor0}{rgb}{0.6902,0.7686,0.8863}
\definecolor{mylcolor1}{rgb}{0.5451,0.8902,0.6941}
\definecolor{mylcolor2}{rgb}{0.9412,0.7490,0.7647}
\definecolor{mylcolor3}{rgb}{0.8627,0.8392,0.9176}
\definecolor{mylcolor4}{rgb}{0.9569,0.9373,0.8667}
\definecolor{mylcolor5}{rgb}{0.7529,0.9020,0.9373}
\definecolor{mylcolor6}{rgb}{0.8750,0.8750,0.8750}

\pgfplotsset{every axis/.append style={grid style={line width=0.6pt,dotted,gray}}}

\begin{document}

\title{Movement Tracking by Optical Flow \\ Assisted Inertial Navigation}

\author{\IEEEauthorblockN{Lassi Meronen}
\IEEEauthorblockA{Aalto University / Saab Finland Oy\\
Espoo, Finland\\
\parbox{.3\textwidth}{\centering lassi.meronen@aalto.fi}}
\and
\IEEEauthorblockN{William J.\ Wilkinson}
\IEEEauthorblockA{Aalto University\\
Espoo, Finland\\
\parbox{.3\textwidth}{\centering william.wilkinson@aalto.fi}}
\and
\IEEEauthorblockN{Arno Solin}
\IEEEauthorblockA{Aalto University\\
Espoo, Finland\\
\parbox{.3\textwidth}{\centering arno.solin@aalto.fi}}
}

\maketitle

\begin{abstract}
Robust and accurate six degree-of-freedom tracking on portable devices remains a challenging problem, especially on small hand-held devices such as smartphones. For improved robustness and accuracy, complementary movement information from an IMU and a camera is often fused. Conventional visual-inertial methods fuse information from IMUs with a sparse cloud of feature points tracked by the device camera. We consider a visually dense approach, where the IMU data is fused with the dense optical flow field estimated from the camera data. Learning-based methods applied to the full image frames can leverage visual cues and global consistency of the flow field to improve the flow estimates. We show how a learning-based optical flow model can be combined with conventional inertial navigation, and how ideas from probabilistic deep learning can aid the robustness of the measurement updates. The practical applicability is demonstrated on real-world data acquired by an iPad in a challenging low-texture environment.
\end{abstract}

\section{Introduction}
\begin{figure}[t!]
  \tiny
  \begin{tikzpicture}[outer sep=0]

    \node[minimum width=\columnwidth,minimum height=0.75\columnwidth,rounded corners=2mm,path picture={
      \node at (path picture bounding box.center){
        \includegraphics[width=\columnwidth]{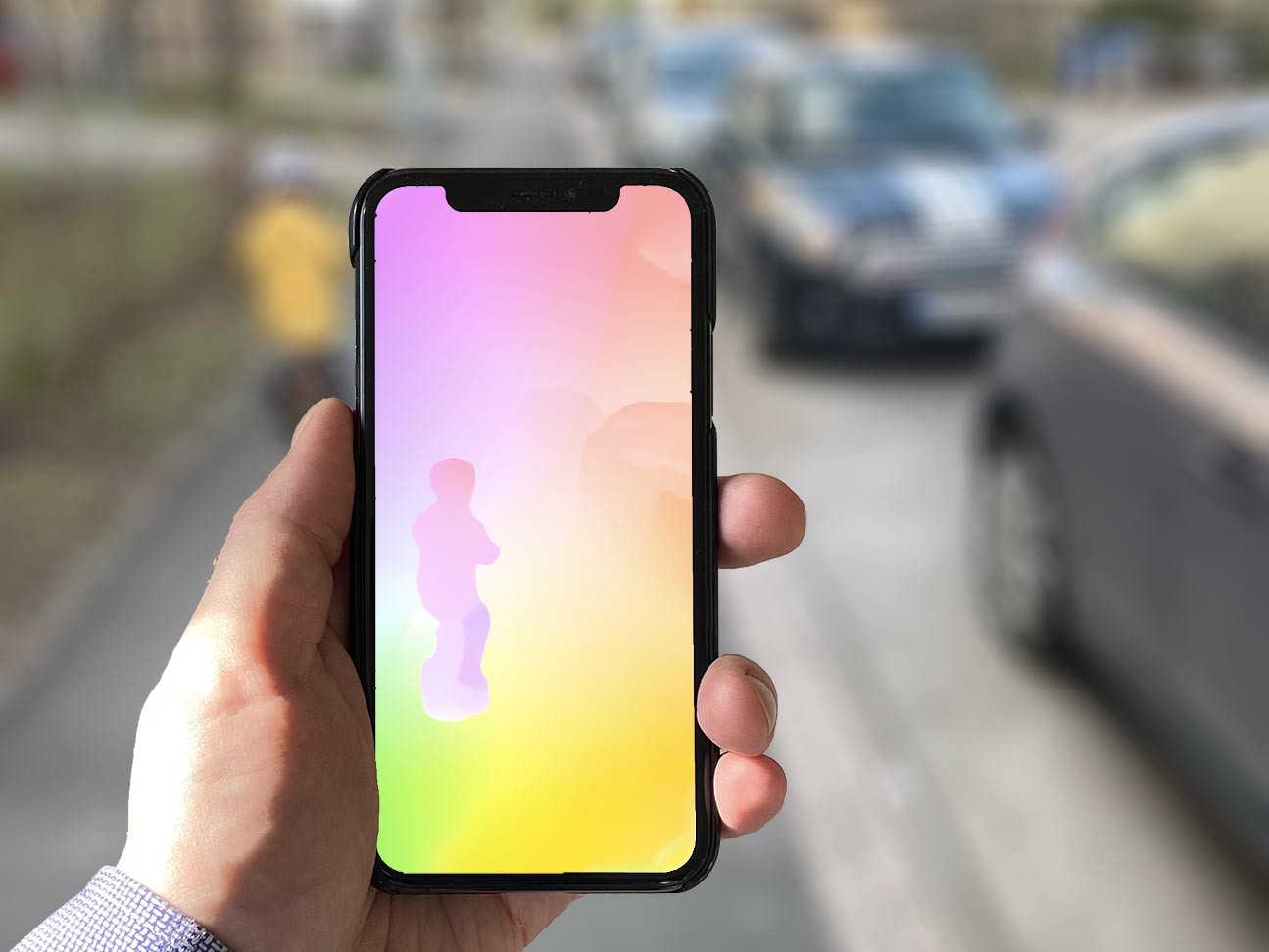}
      };}] at (0,0) {};

    \node[minimum width=2.5cm,circle,path picture={
      \node at (path picture bounding box.center){
        \includegraphics[width=2.5cm]{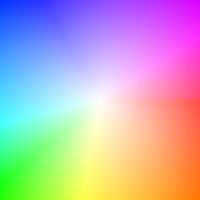}
      };}] at (2.75,-1.85){};

    \draw[draw=black,thick,-latex] (0,-1) -- (0,0);
    \draw[draw=black,thick,-latex] (0,-1) -- (1,-1);
    \draw[draw=black,thick,-latex] (0,-1) -- (.3,-1.3);

    \draw[draw=black,thick,-latex] (0,1.7) -- (1,1.7);
    \draw[draw=black,thick,-latex] (0,1.7) -- (0,0.7);
    \draw[draw=black,thick,-latex] (0,1.7) -- (-.3,2);

    \node[text width=2cm,align=center] at (2.75,-2.75) {Optical flow \\ direction};    
    \node[text width=5mm,align=center] at (-0.4,-1) {IMU frame};            
    \node[text width=5mm,align=center] at (-0.4,1.5) {Camera frame};
    \node[text width=2cm,align=center,execute at begin node=\setlength{\baselineskip}{9pt}] at (-3,2) {\small\it Moving objects cause outliers in the global flow field};
    \node[text width=2cm,align=center,execute at begin node=\setlength{\baselineskip}{9pt}] at (3,1.5) {\small\it Stationary objects in the view agree with the movement predicted by the IMU};
            
  \end{tikzpicture}
  \caption{Outline of the approach. The IMU provides fast-sampled movement information, while the camera is used for estimating the optical flow between camera frames. The optical flow (shown on the phone screen, with the colour wheel showing the flow direection) is used for correcting for inertial drift/learning IMU biases. Non-static objects in the scene are outliers that do not agree with the device movement.}
  \label{fig:teaser}
\end{figure}
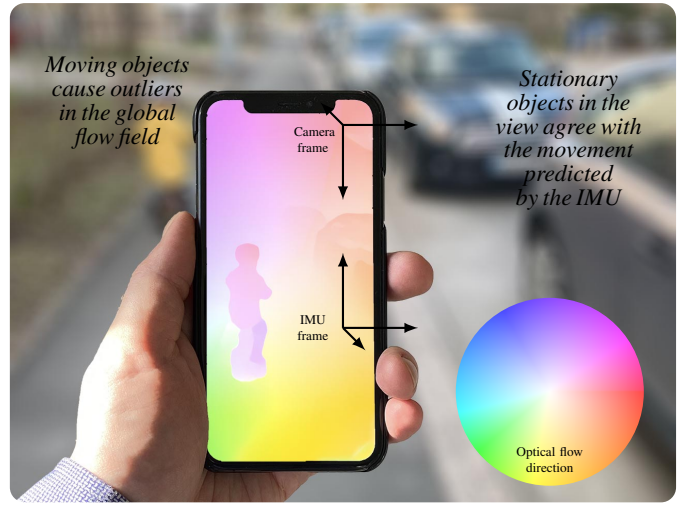

\noindent
We consider the task of motion tracking with inertial sensors under the presence of auxiliary data from a rigidly attached camera in the sensor frame. Our approach takes an orthogonal direction to gold-standard visual-inertial tracking methods (\eg, \cite{Li+Kim+Mourikis:2013, Hesch+Kottas+Bowman+Roumeliotis:2014, Bloesch+Omari+Hutter+Siegwart:2015,Solin+Cortes+Rahtu+Kannala:2018-WACV}), where instead of tracking a set of sparse visual features, we fuse the inertial data with the dense optical flow field that is tracked by a deep neural network. \cref{fig:teaser} shows an example of the optical flow captured on a smartphone.

Inertial navigation (see, \eg, \cite{Jekeli:2001,Bar-Shalom+Li+Kirubarajan:2001,Titterton+Weston:2004,Britting:2010}) is a fundamental and well-studied paradigm for movement tracking. The idea itself can be derived from basic physics: observations of local accelerations in the sensor coordinate frame are rotated by the orientations tracked by a gyroscope (the 3D orientation comes as a by-product), whilst the influence of the Earth's gravity is accounted for. Acceleration can then be turned into position via double-integration.

Inertial navigation with high-grade industrial sensors can be seen as a somewhat solved problem. However, there has been renewed interest \cite{Solin+Cortes+Rahtu+Kannala:2018-FUSION,cortes2018mlsp, Trigoni, Yan+Shan+Furukawa:2018,Solin+Cortes+Kannala:2019-FUSION} in inertial navigation due to the development and popularity of light-weight consumer-grade electronic devices (such as smartphones, tablets, drones, and wearables), which typically include accelerometers and gyroscopes. In this context, inertial navigation is very challenging due to the low quality of consumer-grade inertial sensors. Problems arise due to the double-integration of observed accelerations, which causes rapid error accumulation from high noise-levels in cheap and small inertial measurement units (IMUs) implemented as microelectromechanical systems (MEMS). The problem becomes even more severe due to small inevitable errors in attitude estimation which cause the sensed gravitation to leak into the integrated accelerations.

On these devices inertial navigation can be accomplished through information fusion, where the the estimation is complemented by additional (preferably orthogonal) information sources that can constrain the drift---and/or help the model learn the sensor biases. A prominent approach in this space are the methods derived for foot-mounted consumer-grade inertial sensors \cite{Foxlin:2005,Nilsson+Zachariah+Skog+Handel:2013}. The rationale in attaching the sensor to a foot is that during gait, the foot is at (approximate) standstill for short periods, during which the method can use this information in the form of a zero-velocity update. This makes the inertial navigation problem considerably easier than in the general case since the drift can be constrained (see the OpenShoe project \cite{Nilsson+Zachariah+Skog+Handel:2013, Nilsson+Gupta+Handel:2014}). However, automatic zero-velocity updates are not applicable for handheld or flying devices, where there are no guarantees of regularly repeating moments of standstill.

In the absence of full zero-velocity updates, other information sources can be used. Recent progress in smartphone inertial navigation has largely been due to methods tailored to learning the additive and multiplicate IMU biases online \cite{Solin+Cortes+Rahtu+Kannala:2018-FUSION} by utilizing learnt priors for regressing bounds of speed \cite{cortes2018mlsp} or by constraining the movement by placing priors on the sensible range of instantaneous velocities, by utilising barometric height information, loop-closures, or manual position fixes \cite{Solin+Cortes+Rahtu+Kannala:2018-FUSION, cortes2018advio}. A small number of additional measurements may allow accurate motion trajectory reconstruction by constraining the dynamics, however these additional measurements tend to restrict the generality of the method.

A widely applied approach is to fuse movement from a video camera with the IMU data. Visual-inertial odometry (VIO, \cite{Li+Kim+Mourikis:2013, Hesch+Kottas+Bowman+Roumeliotis:2014, Bloesch+Omari+Hutter+Siegwart:2015,Solin+Cortes+Rahtu+Kannala:2018-WACV}) is one such approach, which is capable of providing full six degree-of-freedom motion tracking in visually distinguishable environments. Fusion of inertial navigation with visual methods has been immensely successful and enabled various use cases, such as autonomous vehicles and augmented reality. This is largely due to the complementary strengths and weaknesses in visual vs.\ inertial navigation: visual methods help prevent large scale drift, but require the environment to be visually rich, the movements not too fast, and are not able to infer absolute scale (in the monocular single-camera case), while inertial methods provide accurate small-scale movement information without restrictions on the environment or movement, but suffer from inertial large scale drift.

The conventional approach to VIO tracks the movement of a sparse set of visually distinctive features (such as corner points) between frames using classical feature extractors, before fusing these with inertial data either in a sparse pose graph (such as in \cite{mur2017visual}) or in a state space framework (such as in \cite{Li+Kim+Mourikis:2013,Solin+Cortes+Rahtu+Kannala:2018-WACV}). We take a different approach, showing proof-of-concept results on how the dense optical flow (see \cite{Dosovitskiy+Fischer+Ilg:2015,Ilg+Mayer+Saikia:2016}) between camera frames can be directly fused with a VIO framework. While attempts have been made at this before, to our knowledge this is the first attempt to do this using learning-based optical flow methods that can account for visual cues in a way that incorporates uncertainty about the optical flow.

The contributions of this paper are the following. \emph{(i)}~We show how principles form Bayesian deep learning can be used to quantify the uncertainty of the optical flow output from a state-of-the-art optical flow estimation network, \emph{(ii)}~we utilise the optical flow to constrain measurements in movement estimation using inertial sensors, and \emph{(iii)}~we show proof-of-concept estimation results on empirical data acquired in a visually challenging environment by a tablet device.

\section{Methods}
\label{sec:methods}
\noindent
Our approach utilises optical flow observations from a camera that is rigidly attached to the IMU in order to assist inertial navigation. In \cref{sec:flow}, we explain how the optical flow is calculated between camera frames by leveraging state-of-the-art learning-based computer vision methods. Next, in \cref{sec:uncertainty} we adjust the model in order to quantify uncertainty in the optical flow estimates, using techniques from Bayesian deep learning. \cref{sec:ins} presents the dynamical model for the inertial navigation framework, and \cref{sec:fusion} finally describes the information fusion and inference approach.

\begin{figure*}
  \captionsetup[subfigure]{justification=centering}
  \newcommand{\colorbar}[2]{$#1$~\protect\includegraphics[width=1cm,height=0.8em]{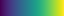}~$#2~\text{px}$}
  \begin{subfigure}[t]{.16\textwidth}
    \sbox0{\includegraphics{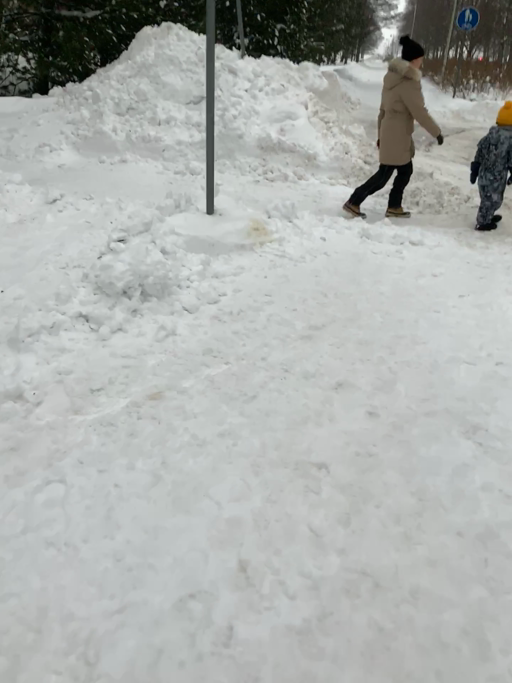}}%
    \includegraphics[width=\textwidth,trim=0 {.0533\wd0} 0 {.0533\wd0},clip]{./fig/frame-02736}
    \caption{Frame \#1}
    \label{fig:frame1}   
  \end{subfigure}
  \hfill
  \begin{subfigure}[t]{.16\textwidth}
    \sbox0{\includegraphics{./fig/frame-02736}}%
    \includegraphics[width=\textwidth,trim=0 {.0533\wd0} 0 {.0533\wd0},clip]{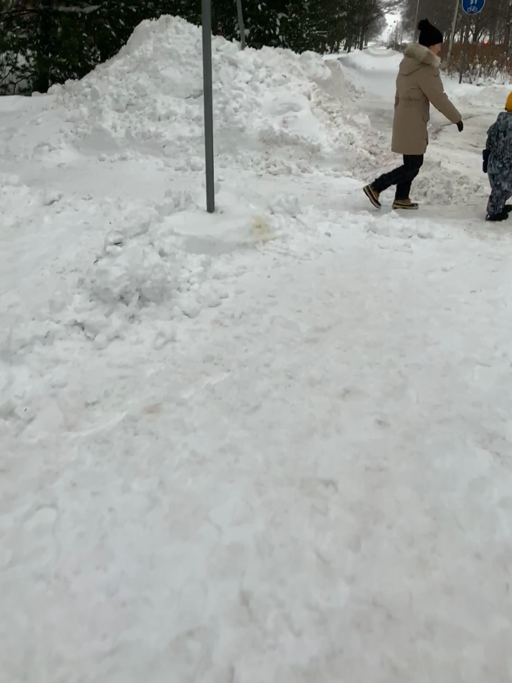}
    \caption{Frame \#2}
    \label{fig:frame2}   
  \end{subfigure}
  \hfill
  \begin{subfigure}[t]{.16\textwidth}
    \sbox0{\includegraphics{./fig/frame-02736}}%
    \includegraphics[width=\textwidth,trim=0 {.0533\wd0} 0 {.0533\wd0},clip]{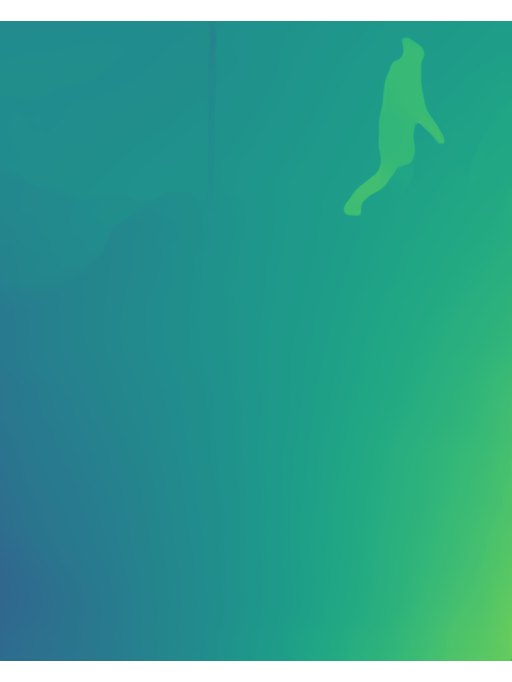}    
    \caption{Horizontal flow \\ \colorbar{-50}{50}}
    \label{fig:flow-u}   
  \end{subfigure}
  \hfill
  \begin{subfigure}[t]{.16\textwidth}
    \sbox0{\includegraphics{./fig/frame-02736}}%
    \includegraphics[width=\textwidth,trim=0 {.0533\wd0} 0 {.0533\wd0},clip]{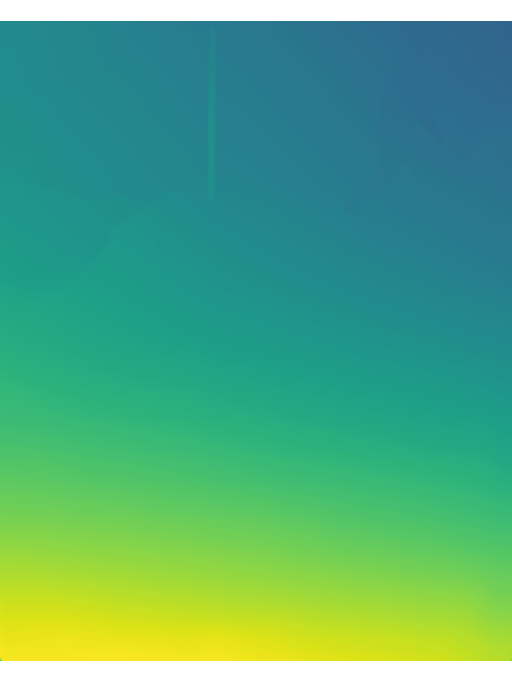}  
    \caption{Vertical flow \\ \colorbar{-50}{50}}
    \label{fig:flow-v}
  \end{subfigure}
  \hfill
  \begin{subfigure}[t]{.16\textwidth}
  	\sbox0{\includegraphics{./fig/frame-02736}}%
    \includegraphics[width=\textwidth,trim=0 {.0533\wd0} 0 {.0533\wd0},clip]{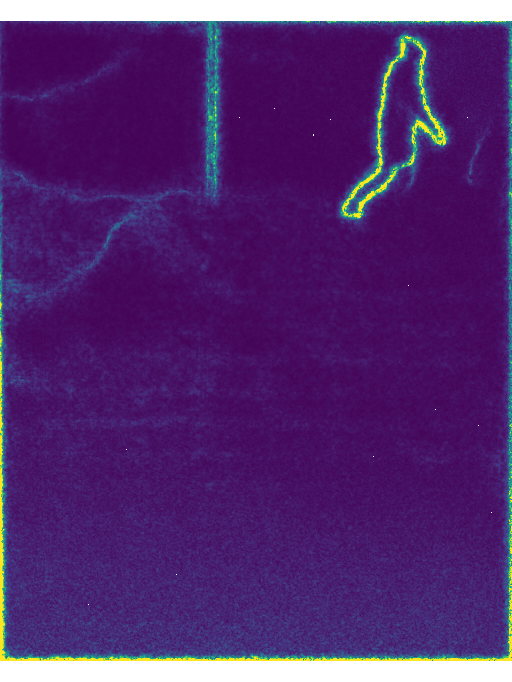} 
    \caption{Uncertainty \#1 \\ \colorbar{0}{1.5}}
    \label{fig:flow-std-1}   
  \end{subfigure}
  \hfill
  \begin{subfigure}[t]{.16\textwidth}
    \sbox0{\includegraphics{./fig/frame-02736}}%
    \includegraphics[width=\textwidth,trim=0 {.0533\wd0} 0 {.0533\wd0},clip]{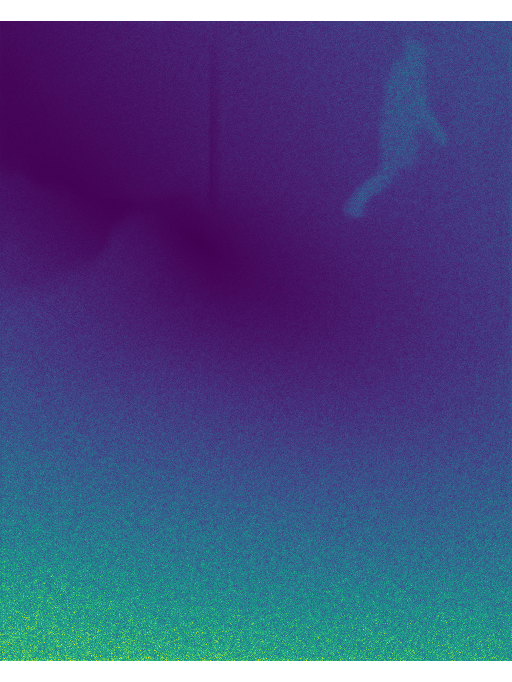} 
        \caption{Uncertainty \#2 \\ \colorbar{0}{15}}
    \label{fig:flow-std-2}
  \end{subfigure}  
  \caption{An example pair of frames from the dataset used in the experiments is visualized in subfigures (a) and (b). The horizontal and vertical optical flows calculated by FlowNet2 are visualized in subfigures (c) and (d) respectively. Subfigures (e) and (f) visualize the flow uncertainty estimates obtained using options \#1 and \#2 from \cref{fig:architecture_fusion} respectively.}
  \label{fig:flow}  
\end{figure*}

\subsection{Optical Flow}
\label{sec:flow}
\noindent
Optical flow (see, \eg, \cite{Hartley+Zisserman:2003,Ilg+Mayer+Saikia:2016}) describes the relative movement of visual objects with respect to an observer. Optical flow arises either from the movement of the observer or movement of the surrounding objects. Since the visual scene appears to the observer as a 2-dimensional projection of a 3-dimensional environment, the optical flow is usually considered on a 2-dimensional projected space, such as an image.

Considering a visual point $\vp$ appearing in a location $(u_1, v_1)$ in a scene at time $t_1$ and in another location $(u_2, v_2)$ at time $t_2$, the optical flow of $\vp$ over time $\Delta t = t_2 - t_1$ is a 2-dimensional vector $\vdelta = (\Delta u, \Delta v)\T$, where $\Delta u = u_2 - u_1$ and $\Delta v = v_2 - v_1$. Usually this optical flow $\vdelta$ is assigned to the location $(u_1, v_1)$ on the projected space, as the original point $\vp$ is unknown. This means that considering a point $(u_1, v_1)$ in a projected visual scene, the optical flow describes how that point will appear to move in the visual scene over time $\Delta t$. Optical flow is a widely applied concept in computer vision, and it can be roughly divided into two approaches: traditional geometric sparse (point-wise/local) methods and learning-based dense methods.

Optical flow is often applied on images of a video, such that two consecutive frames $I_1$ and $I_2$ of the video are analysed and the optical flow for points of $I_1$ is determined to describe how they appear to move in the visual scene transition from $I_1$ to $I_2$. FlowNet \citep{Dosovitskiy+Fischer+Ilg:2015} is a convolutional neural network implementation for computing a dense optical flow for consecutive image frames. The model training is performed in an end-to-end supervised learning setup by training the neural network on synthetic video frames and corresponding ground truth optical flow values. FlowNet~2.0 \citep{Ilg+Mayer+Saikia:2016} is an improved version of the standard FlowNet, constructed by stacking multiple convolutional networks together. 

In \cref{fig:architecture_fusion}, we have extracted the last part of the FlowNet~2.0 neural network structure proposed by Ilg~\etal~\citep{Ilg+Mayer+Saikia:2016}. The network takes in two images and propagates them through three sequential convolutional neural networks. Each of these networks calculates an optical flow estimate, which is then used as an input image for the next one. This way each consecutive network can improve on the results from the previous one. The FlowNet~2.0 architecture also has a separate convolutional neural network for estimating small displacements, which operates in parallel with the rest of the network. Lastly, the outputs from these networks are concatenated and fed into a final convolutional network, the `fusion' network shown in \cref{fig:architecture_fusion}, which fuses the information to generate a final optical flow estimate. \cref{fig:flow} shows a pair of example video frames and their associated optical flow, with vertical and horizontal flow displayed separately.

The impressive performance of FlowNet~2.0 is partly due to its elaborate training schedule. The three stacked convolutional networks are trained sequentially by training one network separately first, then fixing the weights of the first network and stacking the second network after it, and proceeding to train the second network parameters. Each of the training steps consist of initial training on easy data from the Flying Chairs dataset \citep{Dosovitskiy+Fischer+Ilg:2015} followed by training on more challenging data from the FlyingThings3D dataset \citep{Mayer+Ilg+Hausser:2016}. Additionally, the network designed to estimate optical flow of small displacements is trained separately on a dataset called ChairsSDHom \citep{Ilg+Mayer+Saikia:2016}, which is specifically designed for this purpose. The training uses only synthetic data and a trained model can be used on any sequence of images independent of the training procedure.

\begin{figure}[t!]
\begin{tikzpicture}[font=\footnotesize, scale=1,outer sep=0]

\tikzstyle{block}=[shape=rectangle, draw=mycolor0, line width=1, fill=mycolor0!10, rounded corners=1pt]
\tikzstyle{line1}=[->, draw=mycolor1, line width=2, rounded corners=1pt]
\tikzstyle{line2}=[->, draw=mycolor2, line width=2, rounded corners=1pt]
\tikzstyle{line3}=[->, draw=mycolor3, line width=2, rounded corners=1pt]

\node[block](node1) at (0,-0.3) {$11\times N_1\times N_2$};
\node[block](node2) at (0,-1.1) {$64\times N_1\times N_2$};
\draw[line1] (node1) to (node2);
\node[block](node3) at (0,-2) {$128\times \frac{N_1}{2}\times \frac{N_2}{2}$};
\draw[line1] (node2) to (node3);

\node[block](node4) at (2.7,-2) {$128\times \frac{N_1}{4}\times \frac{N_2}{4}$};
\draw[line1] (node3) to (node4);
\node[block](node5) at (2.7,-3.2) {$32\times \frac{N_1}{2}\times \frac{N_2}{2}$};
\draw[line2] (node4) to (node5);
\node[block](node6) at (5.2,-2) {$2\times \frac{N_1}{4}\times \frac{N_2}{4}$};
\draw[line1] (node4) to (node6);
\node[block](node7) at (5.2,-3.2) {$2\times \frac{N_1}{2}\times \frac{N_2}{2}$};
\draw[line2] (node6) to (node7);

\draw [dashed, -, draw=black, line width=2] (-0.5,-2.6) to (4,-2.6);
\draw [dashed, -, draw=black, line width=2] (4,-1) to (4,-2.6);
\node[shape=rectangle, line width=0](label2) at (4,-0.8) {Dropout layer option \#1};

\draw [dashed, -, draw=black, line width=2] (1,-10.1) to (5,-10.1);
\node[shape=rectangle, line width=0](label3) at (-0.5,-10.1) {Dropout layer option \#2};

\node[shape=circle, line3] (con1) at (2,-4.2) {};
\draw[line3] (node5) |- (2,-3.7) |- (con1);
\draw[line3] (node3) |- (con1);
\draw[line3] (node7) |- (con1);

\node[block](node8) at (2,-5) {$162\times \frac{N_1}{2}\times \frac{N_2}{2}$};
\draw[line3] (con1) to (node8);
\node[block](node9) at (2,-6) {$32\times \frac{N_1}{2}\times \frac{N_2}{2}$};
\draw[line1] (node8) to (node9);
\node[block](node10) at (2,-7) {$2\times \frac{N_1}{2}\times \frac{N_2}{2}$};
\draw[line1] (node9) to (node10);
\node[block](node11) at (2,-8) {$2\times N_1\times N_2$};
\draw[line2] (node10) to (node11);

\node[block](node12) at (4.7,-7) {$16\times N_1\times N_2$};
\draw[line2] (node8) -| (node12);

\node[shape=circle,line3](con2) at (3,-9) {};
\draw[line3] (node2) -| (-1.5,-2) |- (con2);
\draw[line3] (node12) |- (con2);
\draw[line3] (node11) |- (3,-8.5) |- (con2);

\node[block] (node13) at (3,-9.7) {$82\times N_1\times N_2$};
\draw[line3] (con2) to (node13);

\node[block](node14) at (3,-10.6) {$16\times N_1\times N_2$};
\draw[line1] (node13) to (node14);

\node[block](node15) at (3,-11.5) {$2\times N_1\times N_2$};
\draw[line1] (node14) to (node15);

\node[shape=rectangle, line width=0](label1) at (0,0.1) {Input};
\node[shape=rectangle, line width=0](label1) at (3,-12) {Final optical flow output};

\end{tikzpicture} 
    \caption{Fusion network architecture for inferring the optical flow (from FlowNet~2.0), showing the two different approaches for applying Monte Carlo dropout for quantifying the variability/uncertainty in the output. The network layers that these dropout options affect are shown with dashed black lines. The fusion network is only the final part of FlowNet~2.0 as seen in Fig.~2 in \citep{Ilg+Mayer+Saikia:2016}. The input to the fusion network has a size of $11 \times N_1 \times N_2$, where $11$ is the number of channels and $N_1 \times N_2$ is the native input image resolution. The 11 channels in the input of the fusion network contain one channel of both large and small displacement flow magnitudes, two channels of both large and small displacement flows, one channel of both large and small displacement brightness errors and three channels of original input frame~1. The numbers inside rectangular blocks represent the number of channels and pixel dimensions of intermediate results within the fusion network. Green (\protect\tikz[anchor=base, baseline]\protect\draw[->, draw=mycolor1, line width=2, yshift=2pt] (0,0) -- (0.7,0);) and red (\protect\tikz[anchor=base, baseline]\protect\draw[->, draw=mycolor2, line width=2, yshift=2pt] (0,0) -- (0.7,0);) arrows represent convolution and de-convolution layers in the network. Purple arrows (\protect\tikz[anchor=base, baseline]\protect\draw[->, draw=mycolor3, line width=2, yshift=2pt] (0,0) -- (0.7,0);) and circles represent channel-wise concatenation. The output has two channels in the native input image resolution, representing the horizontal and vertical optical flows. \cref{fig:flow} shows an example pair of input frames, example horizontal and vertical flow outputs and example uncertainty estimates for both the Monte Carlo dropout options.}
    \label{fig:architecture_fusion}   
\end{figure}
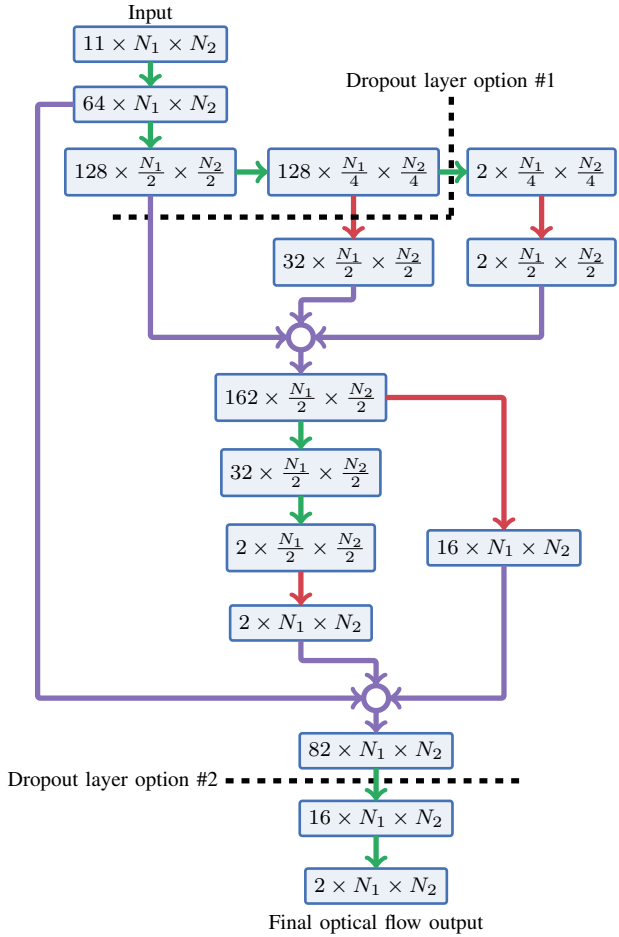

\subsection{Uncertainty Quantification}
\label{sec:uncertainty}
\noindent
Optical flow estimation quality may deteriorate due to abrupt camera movements, which are common with hand-held devices, or due to challenging visual conditions (due to low texture surfaces, varying lighting conditions, or motion blur), such as a snowy environment. Using poor optical flow estimates to correct inertial navigation drift would hinder recovery of the true movement. To account for this issue, we use uncertainty quantification for the calculated optical flow to determine which flow values can be more reliably used for the inertial navigation drift correction. Uncertainty quantification is implemented using Monte Carlo dropout \citep{Gal+Ghahramani:2016}, which involves repeatedly removing a fraction of randomly selected nodes in the network during test time to obtain multiple output predictions for a single pair of input images. $N_\mathrm{MC}$ forward passes are performed for each pair of input images, resulting in $N_\mathrm{MC}$ differing outputs as different randomly selected nodes in the network are deactivated during each forward pass. Optical flow uncertainty can then be estimated by calculating the standard deviation of these $N_\mathrm{MC}$ outputs. This results in an individual uncertainty value for each pixel in the calculated optical flow, separately for the vertical and horizontal directions.

Monte Carlo dropout is applied at one layer in the network, only dropping out nodes in that specific layer. \cref{fig:architecture_fusion} shows two alternative dropout layers that were tested for our application. Examples of uncertainty estimates obtained with both options are visualized in \cref{fig:flow}. Option \#1 lets part of the network bypass the dropout layer unaffected, which leads to borders of visual objects showing high uncertainty, suggesting that the network part affected by dropout layer option \#1 is responsible for adjusting the optical flow at the borders of visual objects. Dropout layer option \#2 applies dropout to a later branch of the fusion network without letting any network branches bypass the layer. By visually inspecting the results, \eg, \cref{fig:flow-std-2}, this option appears to measure general uncertainty with large flow values also having larger standard deviation on average. We adopt option \#2 for uncertainty quantification, as a more general estimate is more suited to inertial navigation drift correction.

\begin{figure*}[!t]
  \pgfplotsset{hide axis,scale only axis,width=\figurewidth,height=\figureheight}
  \begin{subfigure}[b]{.30\textwidth}
    \centering
    \setlength{\figurewidth}{\textwidth}
    \setlength{\figureheight}{\textwidth}  
    \input{./fig/map-gnss.tex}\\[0em]
    \caption{GNSS track}
    \label{fig:map-gnss}
  \end{subfigure}
  \hfill
  \begin{subfigure}[b]{.30\textwidth}
    \centering
    \setlength{\figurewidth}{\textwidth}
    \setlength{\figureheight}{\textwidth}  
    \input{./fig/ARKit_ground_truth.tex}\\[0em]
    \caption{Ground-truth from Apple ARKit}
    \label{fig:map-arkit}    
  \end{subfigure}
  \hfill
  \begin{subfigure}[b]{.30\textwidth}
    \centering
    \setlength{\figurewidth}{\textwidth}
    \setlength{\figureheight}{\textwidth}  
    \input{./fig/RTS_flow+GNSS.tex}\\[0em]
    \caption{Flow-assisted inertial navigation}
    \label{fig:map-ours}  
  \end{subfigure}
  \\[1em] 
  \begin{subfigure}[b]{\textwidth}
    \centering
    \setlength{\figurewidth}{0.0909\textwidth}    
    \begin{tikzpicture}

      \newcommand{\figg}[1]{\includegraphics[width=.95\figurewidth]{./fig/selected_flow_examples/frame-#1.jpg}}
      \foreach \x [count=\i] in {00089,00299,00592,00826,00922,01207,01506,01770,02046,02412,02736}
        \node[text width=.9\figurewidth,align=center,text centered,text depth = 0cm] at ({\figurewidth*\i},0) {\figg{\x}};

      \renewcommand{\figg}[1]{\includegraphics[width=.95\figurewidth]{./fig/selected_flow_examples/#1.jpg}}
      \foreach \x [count=\i] in {88_522, 298_1782, 591_3540, 825_4944, 921_5520, 1206_7230, 1505_9024, 1769_10608, 2045_12264, 2411_14460, 2735_16404}
        \node[text width=.9\figurewidth,align=center,text centered,text depth = 0cm] at ({\figurewidth*\i},-1.28\figurewidth) {\figg{\x}};        

      \foreach \i in {1,...,11}
        \node[text width=5mm,inner sep=0,align=center,text centered,text depth = 0cm] at ({\figurewidth*\i},-1.75\figurewidth) {\textcolor{mycolor2}{\scriptsize\bf \i}};
  
    \end{tikzpicture}
    \caption{The camera views along with the associated optical flow estimates}
    \label{fig:frames}
  \end{subfigure}  
  \begin{minipage}{.1\linewidth}
    \centering
    \begin{tikzpicture}[outer sep=0]

    \node[minimum width=1.5cm,circle,path picture={
      \node at (path picture bounding box.center){
        \includegraphics[width=1.5cm]{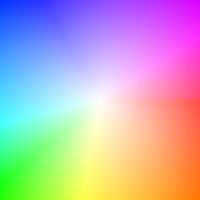}
      };}] at (0,0){};

    \end{tikzpicture}
  \end{minipage}
  \begin{minipage}{.9\linewidth}
    \addtocounter{figure}{-1}
    \captionof{figure}{One example path captured by an Apple iPad. (a)~GNSS/platform location positions with uncertainty radius. (b)~The visual-inertial odometry (Apple ARKit) track that was captured for reference/validation. The ARKit fuses information from the IMU and device camera. The path has been manually aligned to the starting point and orientation. The camera poses for the 11 frames in subfigure d are plotted along the track. (c)~The visual-inertial odometry track for our method combining information from the IMU, device camera in the form of dense optical flow, and GNSS/platform location positions. (d)~Example camera frames and corresponding optical flows from FlowNet~2.0 from the dataset used for the experiments. The visualized optical flow is from between the visualized camera frame and the next frame.}
    \label{fig:maps}
  \end{minipage}
  
\end{figure*}

\subsection{Inertial Navigation}
\label{sec:ins}
\noindent
Our strapdown inertial navigation scheme is based on a state space model with the following state vector:
\begin{equation}
  \vect{x}_k = (\vect{p}_k, \vect{v}_k, \vect{q}_k, \vect{b}_k^\mathrm{a}, \vect{b}_k^{\omega}, \vect{T}_k^\mathrm{a}, \vect{p}_k^-, \vect{q}_k^-),
\end{equation}
where $\vect{p}_k \in \R^3$ represents the position, $\vect{v}_k \in \R^3$ the velocity, and $\vect{q}_k$ the orientation as a unit quaternion at time step $t_k$, $\vect{b}_k^\mathrm{a}$ and $\vect{b}_k^{\omega}$ are, respectively, the additive accelerometer and gyroscope bias components, and $\vect{T}_k^\mathrm{a}$ denotes the diagonal multiplicative scale error of the accelerometer. The state variables $\vect{p}_k^-$ and $\vect{q}_k^-$ denote augmented translation and orientation of the previously seen camera frame that the optical flow will be matched against (see \cref{sec:fusion}).

The dynamical model, $\vect{f}(\vect{x}_k, \vectb{\varepsilon}_k)$, which is a function of the state $\vect{x}_k \sim \N(\vect{m}_{k\mid k},\vect{P}_{k\mid k})$ and the process noise $\vectb{\varepsilon}_k \sim \N(\vectb{0},\vect{Q}_k)$, is given by the mechanization equations (see, \eg, \cite{Titterton+Weston:2004,Nilsson+Zachariah+Skog+Handel:2013} for similar model formulations),

\begin{equation}\label{eq:ins-model}
  \begin{pmatrix}
    \vect{p}_k \\[3pt] \vect{v}_k \\[3pt] \vect{q}_k
  \end{pmatrix}
  =
  \begin{pmatrix}
    \vect{p}_{k-1} + \vect{v}_{k-1}\Delta t_k \\[3pt]
    \vect{v}_{k-1} + [\vect{q}_k \odot (\tilde{\vect{a}}_k + \vectb{\varepsilon}^\mathrm{a}_k) \odot \vect{q}_k^\star - \vect{g}] \Delta t_k \\[3pt]
     \vect{q}_{k-1} \odot \mathscr{Q} \{ (\tilde{\vectb{\omega}}_k + \vectb{\varepsilon}^\omega_k) \Delta t_k \}

  \end{pmatrix},
\end{equation}
where the time step length is given by $\Delta t_k = t_{k} - t_{k-1}$ (note that we {\em do not} assume equidistant sampling times), the accelerometer input is denoted by $\tilde{\vect{a}}_k$ and the gyroscope input by $\tilde{\vectb{\omega}}_k$. Gravity $\vect{g}$ is a constant vector. The symbol $\odot$ denotes quaternion product, and the rotation update is given by the function $\mathscr{Q} \{\vectb{\omega}\}$ from $\R^3$ to $\R^4$ which returns a unit quaternion (see \cite{Titterton+Weston:2004} for details). Note the abuse of notation, where vectors in $\R^3$ are operated along side quaternions in $\R^4$, in the quaternion product, vectors in $\R^3$ are assumed to be quaternions with no real component. The remaining state variables ($\vect{b}_k^\mathrm{a}, \vect{b}_k^{\omega}, \vect{T}_k^\mathrm{a}, \vect{p}_k^-,$ and $\vect{q}_k^-$) have constant dynamics.

We use an Extended Kalman filter (EKF, see, \eg, \cite{Sarkka:2013}) for inference, which approximates the state distributions with Gaussians, $p(\vect{x}_k \mid \vect{y}_{1:k}) \simeq \N(\vect{x}_k \mid \vect{m}_{k\mid k}, \vect{P}_{k\mid k})$, through first-order linearizations. Observations $\vect{y}_{1:k}$ can consist of the optical flow $\vdelta$, sparse visual features, GNSS data, or a combination of these. The dynamics are incorporated into an extended Kalman filter {\em prediction step}:
\begin{equation} \label{eq:prediction_step}
\begin{split}
  \vect{m}_{k \mid k-1} &= \vect{f}(\vect{m}_{k-1 \mid k-1}, \vectb{0}), \\
  \vect{P}_{k \mid k-1} &= \vect{F}_\vect{x} \, \vect{P}_{k-1 \mid k-1} \, \vect{F}_\vect{x}\T + 
   \vect{F}_{\vectb{\varepsilon}} \, \vect{Q}_k \, \vect{F}_{\vectb{\varepsilon}}\T,
\end{split}
\end{equation}
where the dynamical model is evaluated at the state mean from the previous step, and $\vect{F}_\vect{x}$ denotes the Jacobian matrix of $\vect{f}(\cdot, \cdot)$ with respect to $\vect{x}_k$ and $\vect{F}_{\vectb{\varepsilon}}$ denotes the Jacobian with respect to the process noise $\vectb{\varepsilon}_k$, both evaluated at the mean. The process noise covariance is $\vect{Q}_k = \mathrm{blkdiag}(\vectb{\Sigma}^\mathrm{a} \Delta t_k, \vectb{\Sigma}^\omega \Delta t_k)$.

\subsection{Information Fusion}
\label{sec:fusion}
\noindent
We build upon traditional geometric computer vision techniques, leveraging learning-based flow estimation which is well suited to online state inference via non-linear Kalman filtering. In geometric computer vision (\eg, \cite{Hartley+Zisserman:2003}), a camera projection model is characterized by so called {\em extrinsic} (external) and {\em intrinsic} (internal) camera parameters. The extrinsic parameters denote the coordinate system transformations from world coordinates to camera coordinates, while the intrinsic parameters map the camera coordinates to image coordinates. In a {\em pinhole camera} model, this corresponds to
\begin{equation}\label{eq:camera}
  \begin{pmatrix} u & v & 1 \end{pmatrix}\T \propto \MK^\mathrm{cam} \begin{pmatrix} \MR\T & -\MR\T \vp \end{pmatrix} \begin{pmatrix} x & y & z & 1 \end{pmatrix}\T,
\end{equation}
where $(u,v)$ are the image (pixel) coordinates, $\vp_\mathrm{xyz} = (x,y,z) \in \R^3$ are the world coordinates, $\MK^\mathrm{cam}$ is the intrinsic matrix (with camera focal lengths and principal point), and the $\vp \in \R^3$ and  $\MR$ describe the position of the camera centre and the orientation in world coordinates respectively. From \cref{eq:camera}, given a set of fixed world coordinates and a known motion between frames, changes in pixel values $(u,v)$ are driven by the camera pose $P=\{\vp,\MR\}$.

Consider two consecutive frames, $I_1$ and $I_2$, observed by the camera. Given the estimated flow values $\vdelta^{(i)}$ (as described in \cref{sec:flow}, with $i$ referring to a single flow point), we have an estimate for how the pixels $(u_1^{(i)},v_1^{(i)})$ in frame~$I_1$ have moved to pixel locations $(u_2^{(i)},v_2^{(i)})$ in frame~$I_2$:
\begin{equation}\label{eq:flow}
  u_2^{(i)} = u_1^{(i)}+\Delta u^{(i)} \quad \text{and} \quad v_2^{(i)} = v_1^{(i)}+\Delta v^{(i)}.
\end{equation}

We assume the scene to be static, \ie, we assume all optical flow in the images is generated by the movement of the camera rather than objects in the scene. This oversimplification of the real-world setting necessitates the introduction of outlier rejection measures at a later stage. However, assuming the flow comes from movement of the camera implies that it originates from a world coordinate point $\tilde{\vp}_\mathrm{xyz}^{(i)} = (x, y, z, 1)\T$ whose mapping from the world coordinates to the camera coordinates is determined by the relation given in \cref{eq:camera}. This relation is given in terms of state variables of the form
\begin{align}
  \MP_1^\mathrm{cam} &= \MK_1^\mathrm{cam} \begin{pmatrix} \MR(\vq_k^-)\T & -\MR(\vq_k^-)\T \vp_k^- \end{pmatrix}, \\
  \MP_2^\mathrm{cam} &= \MK_2^\mathrm{cam} \begin{pmatrix} \MR(\vq_k)\T & \,\,-\MR(\vq_k)\T \vp_k \end{pmatrix},
\end{align}
(with the rotation also accounting for the IMU-to-camera frame rotation) such that
\begin{equation}\label{eq:flow2}
  \begin{pmatrix} u_1^{(i)} \\ v_1^{(i)} \end{pmatrix} \propto \MP_1^\mathrm{cam}\,\tilde{\vp}_\mathrm{xyz}^{(i)} 
  ~~\text{and}~~  
  \begin{pmatrix} u_2^{(i)} \\ v_2^{(i)} \end{pmatrix} \propto \MP_2^\mathrm{cam}\,\tilde{\vp}_\mathrm{xyz}^{(i)}.
\end{equation}
In geometric computer vision, inferring the unknown world coordinate point $\tilde{\vp}_\mathrm{xyz}$ is a classical triangulation problem which is inherently ill-conditioned. However, applying the the so-called epipolar constraint (see Ch.~12, p.~312 in \cite{Hartley+Zisserman:2003}) to this problem allows for the following formulation (we drop the superscripts $i$ for clarity here)
\begin{equation}\label{eq:triangulation}
  \begin{pmatrix} 
	u_1\,[\MP_1^\mathrm{cam}]_{3,:} - [\MP_1^\mathrm{cam}]_{1,:}\\
	v_1\,[\MP_1^\mathrm{cam}]_{3,:} - [\MP_1^\mathrm{cam}]_{2,:}\\
	u_2\,[\MP_2^\mathrm{cam}]_{3,:} - [\MP_2^\mathrm{cam}]_{1,:}\\
	v_2\,[\MP_2^\mathrm{cam}]_{3,:} - [\MP_2^\mathrm{cam}]_{2,:}\\
  \end{pmatrix}\,\tilde{\vp}_\mathrm{xyz} = \vzero,
\end{equation}
where the notation $[\,\cdot\,]_{j,:}$ extracts the $j$th row of the matrix. This can be solved, for example, by applying a singular value decomposition (SVD, \cite{golub2012matrix}) to the left hand side matrix.

The triangulation problem outlined above requires the optical flow values. Hence, rather than treat the optical flow as the observations, we instead construct a state measurement model that enforces consistency between the model predictions and the flow, subject to additive noise $\vr$. The canonical measurement model for a single point, dropping the time and point indices for convenience, is
\begin{equation}
  \vzero = \vh(\vx, \vdelta+\vr), \quad \vr \sim \N(\vzero, \diag{\sigma_\mathrm{\Delta u}^2, \sigma_\mathrm{\Delta v}^2}),
\end{equation}
where $\vx$ is the state, $\vdelta$ is the observed two-dimensional optical flow for one point, and $\sigma_\mathrm{\Delta u}^2$ and $\sigma_\mathrm{\Delta v}^2$ are the variances of the flow components. Recall from \cref{eq:camera} and subsequently \cref{eq:flow,eq:flow2} how a flow estimate $\bar{\vdelta}$ can be obtained in terms of the camera poses given by the state:
\begin{equation}
  \bar{\vdelta} = 
  \begin{pmatrix} 
	\Delta\bar{u}\\
	\Delta\bar{v}\
  \end{pmatrix} = \MP_2^\mathrm{cam}(\vx)\,\tilde{\vp}_\mathrm{xyz}(\vx,\vdelta) - \MP_1^\mathrm{cam}(\vx)\,\tilde{\vp}_\mathrm{xyz}(\vx,\vdelta),
\end{equation}
where the triangulation outcome $\tilde{\vp}_\mathrm{xyz}$ also depends on the state $\vx$ and the observed flow $\vdelta$ (from \cref{eq:triangulation}). We construct the state observation model $\vh(\cdot,\cdot)$ such that it enforces consistency between our state estimate and the observed flow ($\bar{\vdelta} = \vdelta+\vr$),
\begin{equation}
  \vh(\vx, \vdelta) := \MP_2^\mathrm{cam}(\vx)\,\tilde{\vp}_\mathrm{xyz}(\vx,\vdelta) - \MP_1^\mathrm{cam}(\vx)\,\tilde{\vp}_\mathrm{xyz}(\vx,\vdelta) - \vdelta.
\end{equation}
For an EKF update, we require the Jacobian of this function with respect to the state $\vx$, which gives a $2 \times 26$ matrix and necessitates differentiating over everything in the expression, including the SVD step.

The update step itself follows a standard extended Kalman filter update for each flow point $i$:
\begin{equation}
\begin{split} \label{eq:update}
  \vect{v}_k &= \vzero - \vect{h}(\vect{m}_{k \mid k-1},\vdelta^{(i)}), \\
  \vect{S}_k &= \vect{H}_\vect{x} \, \vect{P}_{k \mid k-1} \, \vect{H}_\vect{x}\T + \vect{H}_\vect{r}\,\vect{R}_k^{(i)}\,\vect{H}_\vect{r}\T, \\
  \vect{K}_k &= \vect{P}_{k \mid k-1} \, \vect{H}_\vect{x}\T \,  \vect{S}_k^{-1}, \\
  \vect{m}_{k \mid k} &= \vect{m}_{k \mid k-1} + \vect{K}_k \, \vect{v}_k, \\
  \vect{P}_{k \mid k} &= \MD_k \, 
\vect{P}_{k \mid k-1} \, \MD_k\T + \vect{K}_k \, \vect{H}_\vect{r}\,\vect{R}_k^{(i)}\,\vect{H}_\vect{r}\T \, \vect{K}_k\T,
\end{split}
\end{equation}
where $\MH_\vx$ denotes the Jacobian of the measurement model $ \vect{h}(\cdot,\cdot)$ with respect to the state variables $\vect{x}$, and $\MH_\vr$ denotes the Jacobian with respect to $\vr$, evaluated at the mean $\vect{m}_{k \mid k-1}$. $\MR_k^{(i)} = \diag{\sigma_\mathrm{i,\Delta u}^2, \sigma_\mathrm{i,\Delta v}^2}$, and $\MD_k = \vect{I}-\vect{K}_k \, \vect{H}_\vect{x}$ (from Joseph's formula, for preserving symmetry/stability during the covariance update). For outlier rejection, it is possible to apply innovation tests or heuristics for accepting updates.

The description above shows how to perform the flow update for a single frame-pair using the current camera pose ($\vp_k$ and $\vq_k$, which are in the state) in relation to the previous camera pose (corresponding to the previous frame) that is also kept in the state ($\vp_k^-$ and $\vq_k^-$). However, in order to keep the state size fixed, we need to replace the augmented pose in the state with the current one after each flow update before moving forward. Following Solin~\etal~\cite{Solin+Cortes+Rahtu+Kannala:2018-WACV}, who do this for a long trail of poses, we can employ the following linear Kalman prediction and update steps for forgetting the past step and remembering the current such that the state covariance structure is properly preserved.

\paragraph{Pose forgetting step}
To remove the current past pose from the end of the state, we apply a linear Kalman prediction step ($\vect{m}_{k \mid k}=\MA_\star\vect{m}_{k \mid k}$, and $\vect{P}_{k \mid k}=\MA_\star\vect{P}_{k \mid k}\MA_\star\T + \MQ_\star$), where the transition matrix discards the pose and the process noise assigns a large variance to the state variables (an uninformative prior): the state transition and process noise covariances are defined as:
\begin{align}
  \MA_\star &= \mathrm{blkdiag}(\MI_{19},\, \vzero_{7}) \quad \text{and} \\
  \MQ_\star &= \mathrm{blkdiag}(\vzero_{19},\, \sigma_{\mathrm{p},0}^2\,\MI_3,\, \sigma_{\mathrm{q},0}^2\,\MI_4).
\end{align}

\paragraph{Pose augmentation step}
To augment the current pose to the end of the state, we `measure' the state variables to be equal, $\MH\,\vx = \vzero$ (that is, $\vp_k^- = \vp_k$ and $\vq_k^- = \vq_k$), by applying a linear Kalman filter update with the following linear measurement model (subject to a small nugget noise term):
\begin{equation}
  \MH_\star = \begin{pmatrix} 
     \MI_3 & \vzero_{3{\times}3} & & \cdots &-\MI_3 & \vzero_{3{\times}4}\\
     \vzero_{4{\times}3} & \vzero_{4{\times}3} & \MI_4  & \cdots & \vzero_{4{\times}3} & -\MI_4
   \end{pmatrix}.
\end{equation}

Processing of the flow updates, the pose forgetting, and the pose augmentation steps for each frame now constrains the inertial navigation system using the optical flow between frame-pairs. This can be interpreted as a fixed-lag smoother. This procedure is needed instead of simply transferring the current state to the end of the state vector in order to maintain relevant covariance information.

\begin{figure*}[!t]
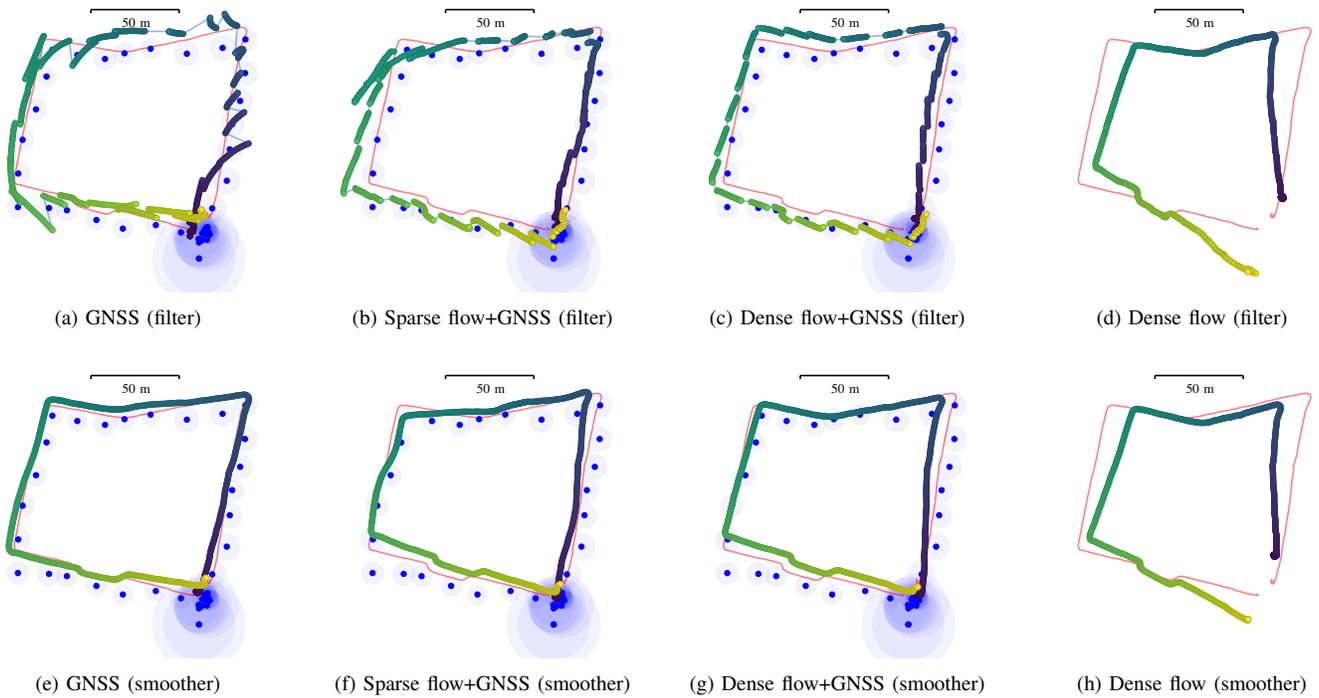

  \pgfplotsset{axis on top,scale only axis,clip=true,ticks=none,axis line style={draw=none}}
  \begin{subfigure}[b]{.22\textwidth}
    \centering
    \setlength{\figurewidth}{\textwidth}
    \setlength{\figureheight}{\textwidth}  
    \input{./fig/EKF_Inertial+GNSS.tex}\\[0em]
    \caption{GNSS (filter)}
    \label{fig:ins-gnss-filter}
  \end{subfigure}
  \hfill
  \begin{subfigure}[b]{.22\textwidth}
    \centering
    \setlength{\figurewidth}{\textwidth}
    \setlength{\figureheight}{\textwidth}  
    \input{./fig/EKF_Inertial+GNSS+sparse.tex}\\[0em]
    \caption{Sparse~flow+GNSS (filter)}
    \label{fig:ins-gnss-sparse-flow-filter}
  \end{subfigure}
  \hfill
  \begin{subfigure}[b]{.22\textwidth}
    \centering
    \setlength{\figurewidth}{\textwidth}
    \setlength{\figureheight}{\textwidth}  
    \input{./fig/EKF_Inertial+GNSS+dense.tex}\\[0em]
    \caption{Dense~flow+GNSS (filter)}
    \label{fig:ins-gnss-flow-filter}
  \end{subfigure}
  \hfill
  \begin{subfigure}[b]{.22\textwidth}
    \centering
    \setlength{\figurewidth}{\textwidth}
    \setlength{\figureheight}{\textwidth}  
    \input{./fig/EKF_Inertial+dense.tex}\\[0em]
    \caption{Dense~flow (filter)}
    \label{fig:ins-flow-filter}
  \end{subfigure}\\[1em]
  \begin{subfigure}[b]{.22\textwidth}
    \centering
    \setlength{\figurewidth}{\textwidth}
    \setlength{\figureheight}{\textwidth}  
    \input{./fig/RTS_Inertial+GNSS.tex}\\[0em]
    \caption{GNSS (smoother)}
    \label{fig:ins-gnss-smoother}
  \end{subfigure}
  \hfill
  \begin{subfigure}[b]{.22\textwidth}
    \centering
    \setlength{\figurewidth}{\textwidth}
    \setlength{\figureheight}{\textwidth}  
    \input{./fig/RTS_Inertial+GNSS+sparse.tex}\\[0em]
    \caption{Sparse~flow+GNSS (smoother)}
    \label{fig:ins-gnss-sparse-flow-smoother}
  \end{subfigure}
  \hfill
  \begin{subfigure}[b]{.22\textwidth}
    \centering
    \setlength{\figurewidth}{\textwidth}
    \setlength{\figureheight}{\textwidth}  
    \input{./fig/RTS_Inertial+GNSS+dense.tex}\\[0em]
    \caption{Dense~flow+GNSS (smoother)}
    \label{fig:ins-gns-flow-smoother}
  \end{subfigure}
  \hfill  
  \begin{subfigure}[b]{.22\textwidth}
    \centering
    \setlength{\figurewidth}{\textwidth}
    \setlength{\figureheight}{\textwidth}  
    \input{./fig/RTS_Inertial+dense.tex}\\[0em]
    \caption{Dense~flow (smoother)}
    \label{fig:ins-flow-smoother}
  \end{subfigure}
  \caption{Filter and smoother track results for different method combinations on the same data. For each method combination, the result track is plotted as coloured points starting at blue color and finishing at yellow colour. In each subfigure the ground truth Apple ARKit track is plotted as a red line. For each method combination that includes the use of GNSS data, the used GNSS positions with their respective uncertainty radiuses are plotted as blue dots surrounded by blue shaded circles. The method combination names for each subfigure represent the included methods as listed in \cref{tbl:results}.}
  \label{fig:results}
\end{figure*}

\section{Experiments}
\label{sec:experiments}
\noindent
The experiments are conducted on a dataset captured with a handheld device while walking along a path in snowy conditions forming a closed loop. The captured data consist of IMU measurements sampled at roughly 100 samples per second, a $1440\times1920$ pixel video captured at 60~fps, and GNSS position measurements with position accuracy information. The experiments compare using inertial navigation with different combinations of additional visual and GNSS information. For each combination, the resulting extended Kalman filter and extended Rauch--Tung--Striebel smoother solutions are calculated. The performance of each method is evaluated via RMSE with respect to a ground truth obtained from the Apple ARKit visual-inertial odometry track that leverages simultaneous localization and mapping. Before the RMSE calculation, the predicted track is rotated and translated to minimize the least-squares distance between the track and the ground-truth (procrustes). This is done to achieve a fair comparison between different methods, as those not using GNSS data have no information regarding absolute location and orientation.

The optical flow for our experiments was calculated using the FlowNet~2.0 Pytorch implementation from NVIDIA \citep{flownet2-pytorch} including a ready trained network. For the flow calculation, the video resolution is reduced to $512\times683$ pixels with a frame rate of 10~fps. To calculate flow uncertainty using Monte Carlo dropout, we use dropout layer option \#2 shown in \cref{fig:architecture_fusion} and perform $N_{MC}=10$ forward passes for calculating flow standard deviation. All methods were implemented and run in Python. The extended Kalman filtering implementation follows the equations shown in \cref{sec:fusion} and additional GNSS measurement updates are performed as standard extended Kalman filter updates. The required Jacobian matrices for the dynamical and measurement models were calculated via automatic differentiation using JAX \citep{jax2018github}. As a baseline comparison for our dense optical flow method, a sparse feature tracking based optical flow was calculated using a Shi--Tomasi corner detector (OpenCV `Good Features to Track') and a pyramidal Lucas--Kanade tracker as is typical in sparse VIO methods (see, \eg, \cite{Li+Kim+Mourikis:2013,Solin+Cortes+Rahtu+Kannala:2018-WACV}). In conventional methods, the pose trail length is typically tens of past poses, while we only consider a pose-pair in the experiments, which makes the estimation challenging.

\subsection{Data Acquisition Setup}
\label{sec:data}
\noindent
We use data from \cite{Solin+Cortes+Kannala:2019-FUSION}. The data set was captured while walking on city streets around a non-square block in Helsinki, Finland. The device used to capture the data was an Apple iPad Pro (11-inch, late-2018 model). Most of the ground was covered in snow and parts of the side walk were completely blocked off by deep snow. The streets along the path are surrounded by 5--10 storey buildings and both parked and moving cars. The capture device was pointed forward and slightly down on the walked path and held at roughly head high. At the start of the recording the device is held against a pole for roughly 5 seconds to ensure no movement. This procedure allows sensor biases to be calibrated online, which is required as these are temperature sensitive. The complete walked path forms a loop of 360 m with a duration of about 300 seconds. \cref{fig:map-arkit} shows the path overlaid on a map and \cref{fig:frames} shows 11 frames captured by the device camera.

To record acceleration, rotation rate, magnetometer readings, barometric pressure, platform locations, video frames, and Apple ARKit poses from the iPad sensors, an app developed for \cite{cortes2018advio} was used. Time synchronization for the multiple data streams is handled by the capture device. The platform locations are GNSS based and are formed by merging satellite and ground based positioning systems. The location measurement quality is reflected in an uncertainty radius associated with every location measurement, as shown in \cref{fig:map-gnss} and \cref{fig:results}. The platform location is recorded in WGS coordinates and transformed to metric ENU (East-North-Up) coordinates to be used in model calculations.

\subsection{Comparisons and Ablation Studies}
\label{sec:comparison}
\noindent
We compare resulting filter (conditioned on data seen up to time $t_k$) and smoother (conditioned on data over the entire track) location tracks for different methods. Each tested method uses the IMU-based dynamical model, but the data used to perform the extended Kalman filter update varies from method to method. \cref{tbl:results} shows RMSE results for different methods, for both the filter and smoother track, listing the different types of measurement update information that each method utilizes. If GNSS measurements are used, the track estimates attempt to align to global geographic coordinates. If they are not used then the orientation and translation are purely relative to starting pose.

\cref{fig:results} shows filter and smoother tracks for four different method combinations from \cref{tbl:results} (the ones with smallest RMSE). As an ablation study, the use of each additional measurement update information is disabled in order to observe how each information source affects the result. Looking at the RMSE values in \cref{tbl:results}, we can see that the best filtering results are obtained using GNSS data and dense optical flow in addition to the IMU information. If GNSS is disabled, only methods utilizing the dense optical flow manage to produce a reasonable track, as seen in the RMSE values and in \cref{fig:ins-flow-filter,fig:ins-flow-smoother}. Significantly, despite these tracks showing some orientation and scale drift, the overall shape is captured even during filtering. The best smoother RMSE value is obtained using only GNSS location data in addition to IMU information. This is explained by the slightly reduced scale of the tracks produced when using the dense optical flow. However, the forward filtering results given by our optical flow updates are superior to those using just GNSS or the sparse flow, suggesting that this approach is better suited to online inference. The top row of \cref{fig:results} shows clear differences in the filtering results, where methods not using the dense flow clearly drift off between the absolute position corrections (\cref{fig:ins-gnss-sparse-flow-filter,fig:ins-gnss-filter}). This large drift is due to poor filter tuning, which is unavoidable with the poor quality of IMU-measurements from the iPad. In \cref{fig:ins-gnss-flow-filter} the GNSS corrects the orientation and translation, whilst the flow prevents inertial drift.

\begin{table}
  \centering
  \caption{Comparison between various information fusion combinations. Each row represents a different combination and the columns indicate which information sources were included with tick marks. \textbf{Methods 4 and 6} represent our novel approach that utilizes dense flow and flow uncertainty, which improves filtering and performs well in the absence of GNSS.}
  \label{tbl:results}
  \begin{tabular*}{\columnwidth}{l c c c c c c c}
    \toprule
     & & & Sparse & Dense & & \multicolumn{2}{c}{RMSE (meters)} \\
    \# & INS & GNSS & flow & flow & Uncertainty & Filter & Smoother \\
    \midrule
    1 & \checkmark & & & & & 2832 & 2838\\
    2 & \checkmark & \checkmark & & & & 8.4 & \textbf{4.8}\\
    3 & \checkmark & & \checkmark & & & 39.3 & 52.4\\
    4 & \checkmark & & & \checkmark & \checkmark & 12.8 & 11.5 \\ 
    5 & \checkmark & \checkmark & \checkmark & & & 10.7 & 6.8\\
    6 & \checkmark & \checkmark & & \checkmark & \checkmark & \textbf{7.1} & 5.8\\

    \bottomrule
  \end{tabular*}  
\end{table}

\section{Discussion and Conclusion}
\noindent
We have presented an approach to visual-inertial odometry in which the full frame information is leveraged (visual cues) through a deep neural network model for optical flow estimation. We complemented FlowNet~2.0---a state-of-the-art learning-based optical flow estimation method---with uncertainty quantification using Monte Carlo dropout, and combined the flow estimates as measurement updates in an inertial navigation system. Our experiments showed that this method improved robustness and stability over traditional sparse visual feature tracking based updates. We argue that this is beneficial in use cases where the visual environment has low texture, direct sunlight, or frames show motion blur, where a dense approach can leverage contextual information across the entire frame. Furthermore, we also showed that our approach can be directly combined with GNSS updates which aligns the odometry result to geographic coordinates. Our approach scales to online operation on a smart device thanks to fast flow calculation by FlowNet~2.0, and with `neural processing units' becoming increasingly popular in handheld devices, we expect information fusion approaches of this kind to become more common.

\section*{Acknowledgements}
\noindent
We acknowledge the computational resources provided by the Aalto Science-IT project and funding from Academy of Finland (324345, 308640) and from Saab Finland Oy.

{\small
\bibliographystyle{ieee}
\bibliography{bibliography}

\begin{thebibliography}{10}\itemsep=-1pt

\bibitem{Bar-Shalom+Li+Kirubarajan:2001}
Y.~Bar-Shalom, X.-R. Li, and T.~Kirubarajan.
\newblock {\em Estimation with Applications to Tracking and Navigation}.
\newblock Wiley-Interscience, New York, 2001.

\bibitem{Bloesch+Omari+Hutter+Siegwart:2015}
M.~Bloesch, S.~Omari, M.~Hutter, and R.~Y. Siegwart.
\newblock Robust visual inertial odometry using a direct {EKF}-based approach.
\newblock In {\em Proceedings of IROS}, pages 298--304, 2015.

\bibitem{jax2018github}
J.~Bradbury, R.~Frostig, P.~Hawkins, M.~J. Johnson, C.~Leary, D.~Maclaurin, and
  S.~Wanderman-Milne.
\newblock {JAX}: {C}omposable transformations of {P}ython+{N}um{P}y programs,
  2018.

\bibitem{Britting:2010}
K.~R. Britting.
\newblock {\em Inertial Navigation Systems Analysis}.
\newblock Wiley-Interscience, New York, 2010.

\bibitem{Trigoni}
C.~Chen, X.~Lu, A.~Markham, and N.~Trigoni.
\newblock {IONet}: Learning to cure the curse of drift in inertial odometry.
\newblock In {\em Proceedings of AAAI}, pages 6468--6476, 2018.

\bibitem{cortes2018mlsp}
S.~Cort{\'e}s, A.~Solin, and J.~Kannala.
\newblock Deep learning based speed estimation for constraining strapdown
  inertial navigation on smartphones.
\newblock In {\em Proceedings of MLSP}, 2018.

\bibitem{cortes2018advio}
S.~Cort{\'e}s, A.~Solin, E.~Rahtu, and J.~Kannala.
\newblock {ADVIO}: {A}n authentic dataset for visual-inertial odometry.
\newblock In {\em Proceedings of ECCV}, pages 419--434, 2018.

\bibitem{Solin+Cortes+Kannala:2019-FUSION}
S.~Cort{\'e}s~Reina, Y.~Hou, J.~Kannala, and A.~Solin.
\newblock Iterative path reconstruction for large-scale inertial navigation on
  smartphones.
\newblock In {\em Proceedings of FUSION}, 2019.

\bibitem{Dosovitskiy+Fischer+Ilg:2015}
A.~Dosovitskiy, P.~Fischer, E.~Ilg, P.~Hausser, C.~Hazirbas, V.~Golkov, P.~Van
  Der~Smagt, D.~Cremers, and T.~Brox.
\newblock Flownet: {L}earning optical flow with convolutional networks.
\newblock In {\em Proceedings of ICCV}, pages 2758--2766, 2015.

\bibitem{Foxlin:2005}
E.~Foxlin.
\newblock Pedestrian tracking with shoe-mounted inertial sensors.
\newblock {\em Computer Graphics and Applications}, 25(6):38--46, 2005.

\bibitem{Gal+Ghahramani:2016}
Y.~Gal and Z.~Ghahramani.
\newblock {Dropout as a Bayesian approximation: representing model uncertainty
  in deep learning}.
\newblock In {\em Proceedings of ICML}, pages 1050--1059, 2016.

\bibitem{golub2012matrix}
G.~H. Golub and C.~F. Van~Loan.
\newblock {\em Matrix Computations}.
\newblock Johns Hopkins University Press, 2012.

\bibitem{Hartley+Zisserman:2003}
R.~Hartley and A.~Zisserman.
\newblock {\em Multiple View Geometry in Computer Vision}.
\newblock Cambridge University Press, Cambridge, UK, 2003.

\bibitem{Hesch+Kottas+Bowman+Roumeliotis:2014}
J.~A. Hesch, D.~G. Kottas, S.~L. Bowman, and S.~I. Roumeliotis.
\newblock Consistency analysis and improvement of vision-aided inertial
  navigation.
\newblock {\em Transactions on Robotics}, 30(1):158--176, 2014.

\bibitem{Ilg+Mayer+Saikia:2016}
E.~Ilg, N.~Mayer, T.~Saikia, M.~Keuper, A.~Dosovitskiy, and T.~Brox.
\newblock {FlowNet 2.0}: {E}volution of optical flow estimation with deep
  networks.
\newblock In {\em Proceedings of CVPR}, pages 2462--2470, 2017.

\bibitem{Jekeli:2001}
C.~Jekeli.
\newblock {\em Inertial Navigation Systems with Geodetic Applications}.
\newblock Walter de Gruyter, Berlin, Germany, 2001.

\bibitem{Li+Kim+Mourikis:2013}
M.~Li, B.~Kim, and A.~Mourikis.
\newblock Real-time motion tracking on a cellphone using inertial sensing and a
  rolling shutter camera.
\newblock In {\em Proceedings of ICRA}, pages 4712--4719, 2013.

\bibitem{Mayer+Ilg+Hausser:2016}
N.~Mayer, E.~Ilg, P.~Hausser, P.~Fischer, D.~Cremers, A.~Dosovitskiy, and
  T.~Brox.
\newblock A large dataset to train convolutional networks for disparity,
  optical flow, and scene flow estimation.
\newblock In {\em Proceedings of CVPR}, pages 4040--4048, 2016.

\bibitem{mur2017visual}
R.~Mur-Artal and J.~D. Tard{\'o}s.
\newblock Visual-inertial monocular {SLAM} with map reuse.
\newblock {\em IEEE Robotics and Automation Letters}, 2(2):796--803, 2017.

\bibitem{Nilsson+Gupta+Handel:2014}
J.-O. Nilsson, A.~K. Gupta, and P.~H\"andel.
\newblock Foot-mounted inertial navigation made easy.
\newblock In {\em Proceedings of IPIN}, pages 24--29, 2014.

\bibitem{Nilsson+Zachariah+Skog+Handel:2013}
J.-O. Nilsson, D.~Zachariah, I.~Skog, and P.~H{\"a}ndel.
\newblock Cooperative localization by dual foot-mounted inertial sensors and
  inter-agent ranging.
\newblock {\em Journal on Advances in Signal Processing}, 2013(1):1--17, 2013.

\bibitem{flownet2-pytorch}
F.~Reda, R.~Pottorff, J.~Barker, and B.~Catanzaro.
\newblock flownet2-pytorch: {P}ytorch implementation of {FlowNet~2.0}:
  {E}volution of optical flow estimation with deep networks.
\newblock \url{https://github.com/NVIDIA/flownet2-pytorch}, 2017.

\bibitem{Sarkka:2013}
S.~S{\"a}rkk{\"a}.
\newblock {\em Bayesian Filtering and Smoothing}.
\newblock Cambridge University Press, 2013.

\bibitem{Solin+Cortes+Rahtu+Kannala:2018-FUSION}
A.~Solin, S.~Cortes, E.~Rahtu, and J.~Kannala.
\newblock Inertial odometry on handheld smartphones.
\newblock In {\em Proceedings of FUSION}, 2018.

\bibitem{Solin+Cortes+Rahtu+Kannala:2018-WACV}
A.~Solin, S.~Cortes, E.~Rahtu, and J.~Kannala.
\newblock {PIVO}: {P}robabilistic inertial-visual odometry for occlusion-robust
  navigation.
\newblock In {\em Proceedings of WACV}, pages 616--625, 2018.

\bibitem{Titterton+Weston:2004}
D.~H. Titterton and J.~L. Weston.
\newblock {\em Strapdown Inertial Navigation Technology}.
\newblock The Institution of Electrical Engineers, 2004.

\bibitem{Yan+Shan+Furukawa:2018}
H.~Yan, Q.~Shan, and Y.~Furukawa.
\newblock {RIDI}: {R}obust {IMU} double integration.
\newblock In {\em Proceedings of ECCV}, 2018.

\end{thebibliography}
}

\end{document}